\documentclass[letterpaper]{article} 
\usepackage{aaai25}  
\usepackage{times}  
\usepackage{helvet}  
\usepackage{courier}  
\usepackage[hyphens]{url}  
\usepackage{graphicx} 
\usepackage{natbib}  
\usepackage{caption} 
\usepackage{amssymb,times}

\usepackage{cite}
\usepackage{verbatim, amsfonts,enumerate}     
\usepackage{bm,array}
\usepackage{graphicx}
\usepackage[cmex10]{amsmath}
\usepackage[font=normalsize]{caption}
\usepackage{algorithm,algorithmic}
\usepackage{color,url}

\usepackage{subcaption} 
\usepackage[table]{xcolor}

\newcommand{\ours}{\texttt{REBEL}}

\title{\LARGE \bf
\texttt{REBEL}: Reward Regularization-Based Approach \\for Robotic Reinforcement Learning from Human Feedback}

\author{
    Souradip Chakraborty\textsuperscript{\rm 1},
    Anukriti Singh\textsuperscript{\rm 1},
    Amisha Bhaskar\textsuperscript{\rm 1},
    Pratap Tokekar\textsuperscript{\rm 1},
    Dinesh Manocha\textsuperscript{\rm 1},
    Amrit Singh Bedi\textsuperscript{\rm 2}
}

\affiliations{
    \textsuperscript{\rm 1}University of Maryland, College Park, MD, USA\\
    \textsuperscript{\rm 2}University of Central Florida, Orlando, FL, USA
}

\usepackage{bibentry}

\begin{document}

\maketitle

\begin{abstract}
The effectiveness of reinforcement learning (RL) agents in continuous control robotics tasks is mainly dependent on the design of the underlying reward function, which is highly prone to reward hacking. A misalignment between the reward function and underlying human preferences (values, social norms) can lead to catastrophic outcomes in the real world especially in the context of robotics for critical decision making. Recent methods aim to mitigate misalignment by learning reward functions from human preferences and subsequently performing policy optimization. However, these methods inadvertently introduce a distribution shift during reward learning due to ignoring the dependence of agent-generated trajectories on the reward learning objective, ultimately resulting in sub-optimal alignment. Hence, in this work, we address this challenge by advocating for the adoption of regularized reward functions that more accurately mirror the intended behaviors of the agent. We propose a novel concept of reward regularization within the robotic RLHF (RL from Human Feedback) framework, which we refer to as \emph{agent preferences}. Our approach uniquely incorporates not just human feedback in the form of preferences but also considers the preferences of the RL agent itself during the reward function learning process. This dual consideration significantly mitigates the issue of distribution shift in RLHF with a computationally tractable algorithm. We provide a theoretical justification for the proposed algorithm by formulating the robotic RLHF problem as a bilevel optimization problem and developing a computationally tractable version of the same. We demonstrate the efficiency of our algorithm {\ours} in several continuous control benchmarks in DeepMind Control Suite \cite{tassa2018deepmind}.

\end{abstract}

\section{Introduction}\label{sec:introduction}
The success of reinforcement learning (RL) relies on the design of an efficient, dense, and accurate reward function for the task at hand \cite{kilinc2021reinforcement,everett2018motion,liu2020robot, terp}. However, in many continuous control robot tasks such as grasping \cite{kilinc2021reinforcement}, motion planning \cite{everett2018motion}, navigation \cite{liu2020robot, terp}, the rewards are naturally sparse \cite{sparse_rwd1, sparse_rwd2, sparse_sc, rauber2021reinforcement, ksd_rl, sparse_5}, thereby making solving them with RL challenging. An effective reward design is necessary. However, in practice, this is often done by trial-and-error or expert-engineering which is inefficient and not always feasible. This problem is compounded when we have multiple attributes in the reward function. For example, in an outdoor navigation scenario, several attributes may significantly affect the reward function, including time taken to reach the goal, distance traveled, risk of collisions, obeying social norms, fuel consumption, etc.~\cite{knox2022reward}.
\begin{figure}[t]
    \centering
    \includegraphics[scale=0.55]{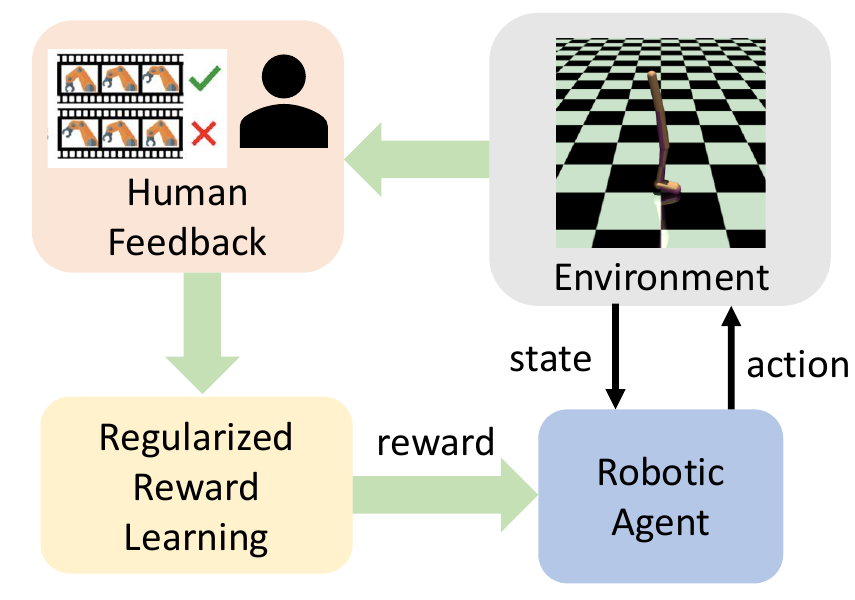}
    \caption{{This figure describes the preference-based robotic reinforcement learning framework from human feedback (RRLHF) \cite{lee2021pebble}. We introduce a novel regularized reward learning into the existing RRLHF framework, which helps  improve the state-of-the-art performance}}
    \label{fig:enter-label}
    \vspace{-5mm}
\end{figure}
It is challenging to hand-design a reward function that can capture multiple attributes. 
Without a principled way to design the reward, we may end up with \textit{reward hacking}, where the agent hacks the reward function and maximizes it without performing the intended task \cite{skalse2022defining}. For instance, let us consider the example of a robotic vacuum cleaner \cite{survey_pref}, where the objective is to clean all dirt and the robot receives a reward when there is no dirt left. Since the robot only perceives dirt through camera images, it might attempt to manipulate what it sees to eliminate the appearance of dirt in its input data (e.g., by going very close to a wall), which would result in a high reward without even accomplishing the intended task of cleaning the dirt.

To address the aforementioned concerns, existing approaches attempt to learn a reward function from expert demonstrations via inverse RL \cite{ziebart_irl, ziebart_cirl} and imitation learning \cite{hussein2017imitation}. The performance of these techniques relies on the quality of the demonstrations collected. Collecting demonstrations can be expensive and sometimes infeasible for large-scale problems \cite{kalashnikov2018qtopt}. Alternatively, there have been efforts to leverage offline data in combination with demonstrations to reduce the dependence on expert feedback. However, such methods require a \emph{coverability} assumption, restricting performance~\cite{xie2022role}.

Recently, preference-based RL has emerged as a compelling framework to solve the sequential decision-making problem by leveraging preferences (feedback) of state-actions pair trajectories \cite{christiano2017deep, lee2021pebble, park2022surf, chakraborty2023parl}. The basic idea is to get pairwise feedback from a user about which of the two trajectories they prefer (can be generalized to $K$-wise) and learn a reward function using Bradley-Terry preference models~\cite{Bradley1952RankAO}. 

The preferences can be at different levels, i.e., trajectory-based, state-based, or action-based. Trajectory-based feedback is the sparsest, and state/action-based feedback can be considered a more dense counterpart. This paradigm is sample-efficient as it does not need humans/experts to provide thousands of high-quality demonstrations but instead provides ranks of the trajectories, which is much easier and more efficient. The most important advantage of the preference-based framework lies in being able to detect and identify the inherent and hidden intentions and motivations from the human/expert preferences and encode them through the reward function. For example, there may be tasks where even the human may not know what the optimal solution looks like but they would still be able to determine which pair is better.

While preference-based RL holds a lot of promise, it also introduces a distribution shift for reward learning because of different data collection policies. More specifically, the dependence of the agent policy generated trajectories are not properly considered while learning the reward function leading to the distribution shift and sub-optimal alignment (also highlighted in \cite{chakraborty2023parl}). One possible way of avoiding this issue is to design a suitable regularization when learning with preferences by incorporating the dependence of the policy in the objective. Hence, one of our main contributions is to design such a regularization term in a principled manner motivated by the Bilevel optimization literature.

More specifically, our \textbf{main contributions} are as follows.

\begin{itemize}
    \item We propose a framework called \textbf{\ours} (\textbf{R}eward r\textbf{E}gularization \textbf{B}ased robotic r\textbf{E}inforcement \textbf{L}earning from human feedback). We introduce a new regularization, termed as \emph{agent preference}, which is given by the value function evaluated at the optimal policy (cf. Sec. \ref{proposed}). This regularization term helps recover the true underlying reward from the human feedback and mitigate the distribution shift in reward learning. 
    \item We show the efficacy of the proposed {\ours} approach on several continuous control locomotion tasks from DeepMind Control Suite \cite{tassa2018deepmind}. We show that in terms of episodic reward return, {\ours} achieves improvement in sample efficiency compared to state-of-the-art baseline methods such as PEBBLE \cite{lee2021pebble} and PEBBLE+SURF \cite{park2022surf}. 
    \item  We also provide a theoretical justification of our proposed regularization method by connecting to first-order Bilevel optimization \cite{ye2022bome} and RLHF, thereby preventing reward hacking.
\end{itemize}

\section{Related Works}\label{sec:lit_survey}
In this section, we discuss the research around reward design in RL specifically emphasizing robotics problems. 
\vspace{2mm}

\noindent \textbf{Reward Shaping and Design.} Sparse rewards in RL create significant challenges in exploration, making it difficult for agents to learn effective policies. Reward shaping offers a practical solution by augmenting sparse extrinsic rewards with intrinsic curiosity terms, a concept originally proposed by \cite{mataric1994reward} and further developed in subsequent studies \cite{reward_1,reward_2,reward_3,reward_4, ksd_rl}. Despite its effectiveness, this approach introduces additional complexity into the MDP, which can lead to suboptimal policies. To address these issues, information-directed RL \cite{steering, hao2022regret} offers a theoretically sound alternative but struggles with scalability in large-scale robotics. IRL, on the other hand, provides a data-driven approach to reward design by inferring reward functions from expert demonstrations, as seen in the influential Max Entropy IRL framework \cite{ziebart_irl, ziebart_cirl}, which robustly handles noisy, imperfect behavior. However, the reliance on high-quality demonstrations remains a critical bottleneck, which recent advancements such as trajectory ranking in suboptimal demonstrations by \cite{brown2019extrapolating} have begun to address.
\vspace{1.5mm}

\noindent {\textbf{Preference-based RL for Robotics.}} Preference-based learning in RL offers a sample-efficient framework for inferring reward functions that align with human or expert preferences. Early work in this area, such as by (Wilson, Fern, and Tadepalli 2012), introduced a Bayesian framework to directly learn a posterior distribution over policies from preferences. Alternatively, (F¨urnkranz et al. 2012) proposed learning a utility function from preferences, maximizing the likelihood of these preferences to improve policy performance. Dueling bandits provide another intriguing approach, focusing on minimizing regret through pairwise action comparisons, though most research here remains theoretical and challenging to scale to large-scale robotics (Dudík et al. 2015; Bengs et al. 2021; Lekang and Lamperski 2019; Pacchiano, Saha, and Lee 2023). Practical approaches like (Christiano et al. 2017) have scaled preference-based learning to large-scale continuous control tasks, but struggled with sample inefficiency due to on-policy learning—a limitation later mitigated by additional demonstrations (Ibarz et al. 2018) and non-binary rankings (Cao, Wong, and Lin 2020). Recently, Pebble (Lee, Smith, and Abbeel 2021) improved efficiency by leveraging off-policy learning and pre-training. However, a significant unresolved issue across these methods is the risk of reward model over-optimization and overfitting, which can undermine policy performance in complex environments.

\section{Problem Formulation}\label{sec:prob}

\subsection{Preliminaries: Policy Optimization in RL}
Let us start by considering the Markov Decision Process (MDP) tuple $\mathcal{M}:=\{\mathcal{S}, \mathcal{A}, \gamma, \mathbb{P}, r\}$, which is a tuple consisting of a state space $\mathcal{S}$, action space $\mathcal{A}$, transition dynamics $\mathbb{P}$, discount factor $\gamma\in(0,1)$, and reward $r:\mathcal{S}\times\mathcal{A} \rightarrow \mathbb{R}$. Starting from a given state $s\in\mathcal{S}$, an agent selects an action $a$ and transitions to another $s^\prime\sim\mathbb{P}(\cdot \mid s,a)$. We consider a stochastic policy that maps states to distributions over actions $\pi_{\theta}: \mathcal{S} \rightarrow \mathbb{P}(\mathcal{A})$, which is parameterized by a parameter vector $\theta\in\mathbb{R}^d$. Hence, we can write the standard finite horizon policy optimization problem as 
\begin{align}\label{main_value_function_Standard}
    \max_{\theta}  V_{s}(\theta):= \mathbb{E}\left[\sum_{h=0}^{H-1} \gamma^h r(s_h, a_h)~|, s_0=s \right],
\end{align}
where the expectation is with respect to the stochasticity in the policy $\pi_{\theta}$ and the transition dynamics $\mathbb{P}$. In \eqref{main_value_function_Standard}, we used notation $V_{s}(\theta)$ for value function in state $s$ to emphasize the dependence on parameters $\theta$. We note that for a given reward function $r$, we can solve the optimization problem in \eqref{main_value_function_Standard} to obtain the optimal policy parameters \cite{schulman2017proximal}. However, establishing a reward function that perfectly aligns with human preferences is a challenging task and might result in potentially misaligned RL policies. There are different ways to learn reward functions as discussed in the introduction (Sec. \ref{sec:introduction}) but we consider the human feedback-based reward learning from preference as follows.

\subsection{Robotic Reinforcement Learning From Human Feedback (RRLHF)} In this subsection, we introduce the problem of reinforcement learning from human feedback for robotic tasks. Following the description mentioned in \cite{lee2021pebble, christiano2017deep}, the {RRLHF} framework can typically be described in three phases as follows. 
\begin{enumerate}
    \item \textbf{Pre-training:} RRLHF starts with a pre-trained policy for the underlying robotic task in an unsupervised manner. We can achieve that via training a policy to maximize the entropy to be able to collect different experiences in the environment \cite{lee2021pebble}. Let's assume the corresponding pre-trained policy as $\pi_0$.
    \item \textbf{Reward Learning from Feedback:} In the second phase, we collect the trajectories by interacting with the training environment and collecting human feedback. Let us call the new reward after feedback as $\hat{r}$. 
    \item \textbf{Policy update}: Once we have a reward learned from the human feedback, then we follow the standard policy optimization to solve the problem \ref{main_value_function_Standard} to learn an updated policy for the new reward function coming from phase 2 as follows
\begin{align}\label{main_value_function_Standard2}
    \max_{\theta}  \mathbb{E}\left[\sum_{h=0}^{H-1} \gamma^h \hat r(s_h, a_h)~|, s_0=s \right].
\end{align}
\end{enumerate}
\textbf{Limitations of Phase 2:} Before discussing the limitations, let us first discuss in detail the existing reward learning procedure. A preference-based feedback is utilized where we collect preferences from humans for each generated trajectory pair ($\tau_i, \tau_j$) generated by a policy $\pi$ and the feedback is represented by $y = [y_i, y_j]$ where $y_i = 1, y_j = 0$ if $\tau_i > \tau_j$ and vice versa. With these preferences, the reward model is learned using the Bradley-Terry (BT) choice model \cite{Bradley1952RankAO} and is mathematically expressed as 
    \begin{align}\label{brad_terry}
    P_{\nu}(\tau_i > \tau_j) = \frac{\exp{G_{\nu} (\tau_i)}}{\exp{G_{\nu} (\tau_i)} + \exp{G_{\nu} (\tau_j)}}
    \end{align}
    where, $G_{\nu} (\tau_i):=\sum_{h=0}^{H-1} \gamma^h r_\nu(s_h^i, a_h^i)$, $\nu$ is the reward model parameter, $(s_h^i, a_h^i)$ denotes state action pairs from the trajectory $\tau_i$,  $P_{\nu}(\tau_i > \tau_j)$ represents the probability of the trajectory $\tau_i > \tau_j$ which models the human preference. Thus the objective of reward learning can be formulated as maximizing the likelihood of human preferences to learn the optimal parameter $\nu$ as
    \begin{align}\label{mle}
        \max_{\nu} \mathbb{E}_{\tau_i, \tau_j \sim \mathcal{D}} [y_i \log P_{\nu}(\tau_i > \tau_j) + y_j \log P_{\nu}(\tau_i < \tau_j)], 
    \end{align}
    %

    \noindent where the expectation is over the trajectories we have in the feedback dataset $\mathcal{D}$. The above-mentioned framework from human feedback is popular but still has not been able to achieve the optimal performance in terms of learning optimal reward, which is clear from the experimental results presented in \cite{lee2021pebble}. There is a gap between the performance of rewards learned from human preferences and the underlying oracle reward (see Figure \ref{fig:enter-label}). In this work, we are interested in investigating this issue and propose improvements.  An important point to note here is that the data set $\mathcal{D}$ is not independent of the policy which we update in phase 2. Since in the RRLHF framework, we repeat the steps in Phase 2 and Phase 3, we need to account for the dependency that the eventual data collection is going to happen at the current optimal policy. This is currently missing from the existing literature, and we propose to fix is by considering a regularization term in the reward learning phase. We highlight the issue with more details in the next section.

\subsection{Key Issue in Reward Learning - Ignoring the Agent Preference \& Distribution Shift}
The reward learning phase in the RRLHF framework is where we utilize preference-based human feedback to design a good reward model to be used in phase 3 of policy learning. It is important to note that in practical implementations such as in \cite{christiano2017deep, lee2021pebble, park2022surf}, phase 2 and phase 3 are repeated in a sequential manner such as from iteration $k=1$ to $K$ and reward and policy are updated for each $k$. At each iteration $k$, we collect data  by interacting with the environment with policy $\pi_{\theta_k^*}$ at each $k$, then we learn a reward model $\nu_{k+1}^*$ by solving \eqref{mle}, followed by the policy update by solving \eqref{main_value_function_Standard2} to obtain $\pi_{\theta_{k+1}^*}$. In the above procedure, we note that when we solve for the policy, we consider the updated value of reward for the value function maximization problem of \eqref{mle}. But on the other hand, when we optimize for reward in  \eqref{main_value_function_Standard2}, we do not pay attention to the current policy $\pi_{\theta_k^*}$. More specifically, we ignore the dependence of the trajectories generated by the policy $\pi_{\theta_k^*}$ while learning the reward model for the next iteration leading to a distribution shift while learning. In other words, prior methods ignore the dependence on the agent policy while learning the reward function which is extremely crucial and plays an important role in the sub-optimal behavior of approaches such as PEBBLE in \cite{lee2021pebble}.

\section{Proposed Approach: REBEL Algorithm} \label{explanation}
In this section, we discuss the theoretical motivation behind agent preference regularization in solving the distribution shift issue prevalent in online RLHF for robotics. In order to do that, we first mathematically express the RLHF problem as a bilevel optimization objective (first highlighted in \cite{chakraborty2023parl}) as
\begin{align}\label{bilevel_rlhf}
    &\ \ \ \ \max_{\nu }  \ \   \mathbb{E}_{y, \tau_0, \tau_1 \sim \rho_h(\tau ; \theta^*(\nu))} [L_{\nu} (y, \tau_0, \tau_1)]
    \\
    \nonumber
    &\text{s.t.} \ \ \theta^*(\nu) := \arg\max_{\theta} \mathbb{E}\left[\sum_{h=0}^{H_{\ell}-1} \gamma^h r_\nu(s_h, a_h)~|, s_0=s \right],
\end{align}
where $L_{\nu} (y, \tau_0, \tau_1)$ represents the likelihood maximization objective given as $L_{\nu} (y, \tau_0, \tau_1) = y \log P_{\nu}(\tau_0 > \tau_1) + (1-y) \log P_{\nu}(\tau_0 < \tau_1)$, as defined in Equation \eqref{mle}. Note that the outer objective requires trajectory samples by solving the inner policy optimization objective and thus, the two objectives cannot be disentangled as done in several past works, including \cite{lee2021pebble, park2022surf} which leads to sub-optimal alignment. The objective intuitively explains that while optimizing the outer objective for learning the reward model, it is important to consider the agent's performance, which is an important factor ignored in prior research leading to distribution shift and sub-optimal alignment.

\vspace{2mm}
\noindent
\textbf{Computational Challenge of Bilevel RLHF}: It is extremely important to note that solving the above bilevel optimization problem is computationally expensive as it requires the estimation of second-order information including hessian and mixed-jacobian terms as shown in \cite{chakraborty2023parl} (refer equation no). For example, in large-scale robotic environments such as DeepMind Control Suite \cite{tassa2018deepmind} and MetaWorld \cite{yu2021metaworld}, the number of parameters for learning the policy gradient are typically in the order of $O(1024 \times k)$, with $k$ being the number of layers. Thus to compute the above hyper-gradient \cite{chakraborty2023parl} with a vanilla network with $k=3$ layers, it requires inverting an hessian matrix of million parameters which eventually leads to an $O(n^3)$ operation which becomes computationally infeasible. Thus, we propose a computationally efficient reformulation of the above objective which requires only first-order information, for the first time in the context of RLHF.

\vspace{2mm}
\noindent
\textbf{Computationally Efficient First-order Reformulation of Bilevel RLHF}: Next, we propose a novel first-order reformulation of the Bilevel RLHF objective to improve its computational traceability for large-scale robotics applications. With the value function-based reformulation, the bilevel objective from Equation \eqref{bilevel_rlhf} can be written as a constrained optimization problem without any approximation as
\begin{align}\label{bilevel_rlhf_fo_0}
    \max_{\nu, \theta}   \mathbb{E}_{y, \tau_0, \tau_1 \sim \rho_h(\tau ; \theta)} [L_{\nu} (y, \tau_0, \tau_1)] \\ \nonumber
    \texttt{s.t} \hspace{3mm} V(\pi_{\theta})  \geq V(\pi_{\theta^*(\nu)}),
\end{align}
where
$V(\pi_{\theta^*(\nu)}) = \mathbb{E}\left[\sum_{h=0}^{H-1} \gamma^h r_{\nu}(s_h, a_h)~|, s_0=s\right]$   with $a_h\sim \pi_{\theta^*(\nu)}$,  $V(\pi_{\theta}) = \mathbb{E}\left[\sum_{h=0}^{H-1} \gamma^h r_{\nu}(s_h, a_h)~|, s_0=s\right]$ with $a_h\sim \pi_{\theta}$, and $V(\pi_{\theta^*(\nu)}) := \max_{\theta} V_{\nu}(\pi_{\theta})$. We note that the constraint in \eqref{bilevel_rlhf_fo_0} satisfies when $\pi_{\theta}$ is the optimal $\pi_{\theta^*(\nu)}$ under the reward function $r_{\nu}$. The primary advantage of this method lies in the fact that it completely relies on the first-order gradient information without requiring the estimation of hypergradient, which can be mathematically shown as
\begin{align}
    \nabla_{\nu} V(\pi_{\theta^*(\nu)}) &= \nabla_1 V(\pi_{\theta^*(\nu)}) + \nabla_{\nu} \pi_{\theta^*(\nu)} \nabla_2 V(\pi_{\theta^*(\nu)})\nonumber \\ 
    & = \nabla_1 V(\pi_{\theta^*(\nu)}),
\end{align}
where the second equality is due to the optimality of the inner objective, since $\nabla_2 V(\pi_{\theta^*(\nu)}) = 0$. This justifies the efficiency in the first-order reformulation of the above objective. Under this setting, the constrained optimization problem can be solved using the penalty-based method as 
\begin{align}\label{bilevel_rlhf_fo}
    \max_{\nu, \theta}   \mathbb{E}_{y, \tau_0, \tau_1 \sim \rho_h(\tau ; \theta)} &[L_{\nu} (y, \tau_0, \tau_1)] \nonumber
    \\
    &+ \lambda (V(\pi_{\theta}) - V(\pi_{\theta^*(\nu)})),
\end{align}
where $\lambda \geq 0$ represents the penalty coefficient.

\vspace{2mm}
\noindent
\textbf{Improved Computational Tractability with Lower bound:}  We note that in order to further improve the computational traceability of our proposed algorithm, we can further simplify the above objective \eqref{bilevel_rlhf_fo}. To do that, we instead of maximizing the first-order objective defined in the equation. \eqref{bilevel_rlhf_fo}, rather we maximize a lower bound of the above objective, where we know that without loss of generality, $V_{\nu}(\pi_{\theta}) = \mathbb{E}\left[\sum_{h=0}^{H-1} \gamma^h r_{\nu}(s_h, a_h)~|, s_0=s, a_h\sim \pi_{\theta}\right]$ satisfies under the condition $V_{\nu}(\pi_{\theta}) \geq 0$ because $r_{\nu}(s, a)\geq 0$ without loss of generality. Thus under this condition, we can maximize the lower bound of the objective defined in equation \eqref{bilevel_rlhf_fo} as
\begin{align}\label{bilevel_rlhf_fo_lb}
    \max_{\nu, \theta}   \mathbb{E}_{y, \tau_0, \tau_1 \sim \rho_h(\tau ; \theta)} [L_{\nu} (y, \tau_0, \tau_1)]
    - \lambda  V(\pi_{\theta^*(\nu)})
\end{align}
where this is the maximization of the lower bound of the objective \eqref{bilevel_rlhf_fo}. The above relaxation and maximizing the lower-bound helps us in designing an extremely efficient and computationally tractable first-order algorithm described in the next section. 

\section{Proposed Algorithm and Approach}\label{sec:proposed}
In this section, we provide a detailed description of the algorithmic design derived from the lower bound maximization objective in equation \eqref{bilevel_rlhf_fo_lb}. We also provide an intuitive explaination of the value-regularized objective via connecting it to agent preference for reward-regularized RLHF.

\begin{algorithm}[t!]
\caption{\textbf{\ours:} Reward Regularization Based Robotic Reinforcement Learning from Human Feedback}
\begin{algorithmic}[1]
\STATE \textbf{Initialize:} Policy parameters $\pi_\theta$, reward function $r_\nu$; preference dataset $\mathcal{D} \leftarrow \emptyset$; replay buffer $\mathcal{B} \leftarrow \emptyset$.
\STATE \textbf{Set hyperparameters:} Learning rate $\eta$, discount factor $\gamma$, regularization coefficient $\lambda$, number of reward learning steps $k_1$, number of trajectories $M$.
\FOR{iteration $1$ to $T$}
    \STATE \textbf{Collect and update preference dataset:} Using the current policy $\pi_\theta$ to collect $M$ trajectory pairs $\{(\tau_0, \tau_1, y)_j\}_{j=1}^M$, where $y$ represents human feedback labels. Then update the dataset $\mathcal{D} \leftarrow \mathcal{D} \cup \{(\tau_0, \tau_1, y)_j\}_{j=1}^M$
    \STATE \FOR{each \textbf{reward learning} step $1$ to $k_1$}
        \STATE sample a batch $\{(\tau_0, \tau_1, y)_j\}_{j=1}^D \sim \mathcal{D}$
        \STATE update reward parameters $\nu$ by minimizing \eqref{bilevel_rlhf_fo_lb} with respect to $\nu$
    \ENDFOR
    \STATE \textbf{Policy update:} Interact with the environment using the current policy $a_t \sim \pi_\theta(a_t | s_t)$ to collect transitions $(s_t, a_t, s_{t+1}, r_\nu(s_t, a_t))$.
    \STATE Update: $\mathcal{B} \leftarrow \mathcal{B} \cup \{(s_t, a_t, s_{t+1}, r_\nu(s_t, a_t))\}$.
    \STATE Sample batch  $\{(s_t, a_t, s_{t+1}, r_\nu)\}_{j=1}^B \sim \mathcal{B}$.
    \STATE Update the policy $\pi_\theta$ with policy gradient in \eqref{bilevel_rlhf_fo_lb} with respect to $\theta$
\ENDFOR
\end{algorithmic}
\end{algorithm}

\begin{figure*}[ht]
    \centering
        \begin{minipage}{0.32\textwidth}
        \centering
        \includegraphics[width=\linewidth]{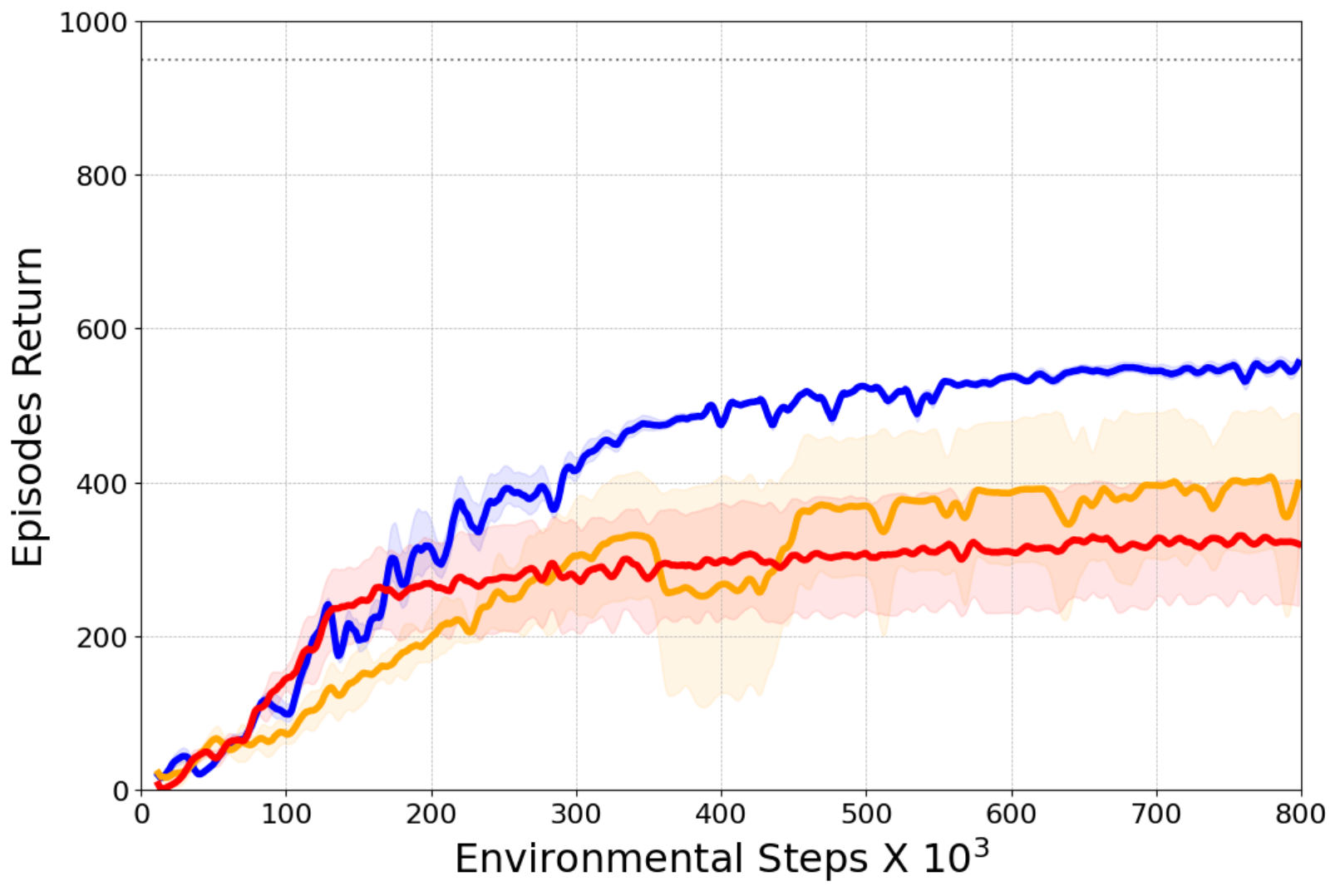}\\
        (a) Cheetah
    \end{minipage}
    \begin{minipage}{0.32\textwidth}
        \centering
        \includegraphics[width=\linewidth]{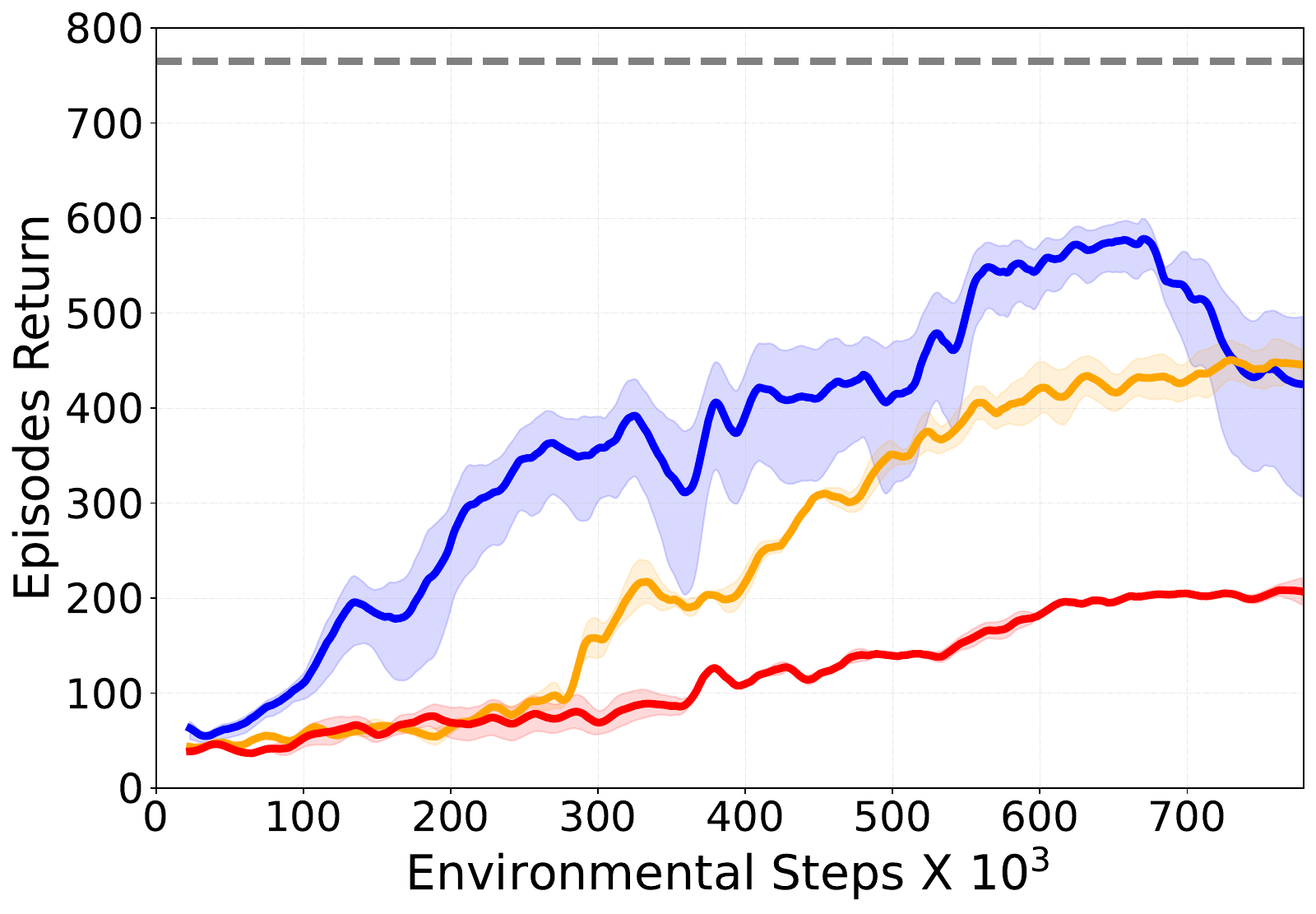}\\
        (a) Visual Walker (Pixel-based)
    \end{minipage}
    \begin{minipage}{0.32\textwidth}
        \centering
        \includegraphics[width=\linewidth]{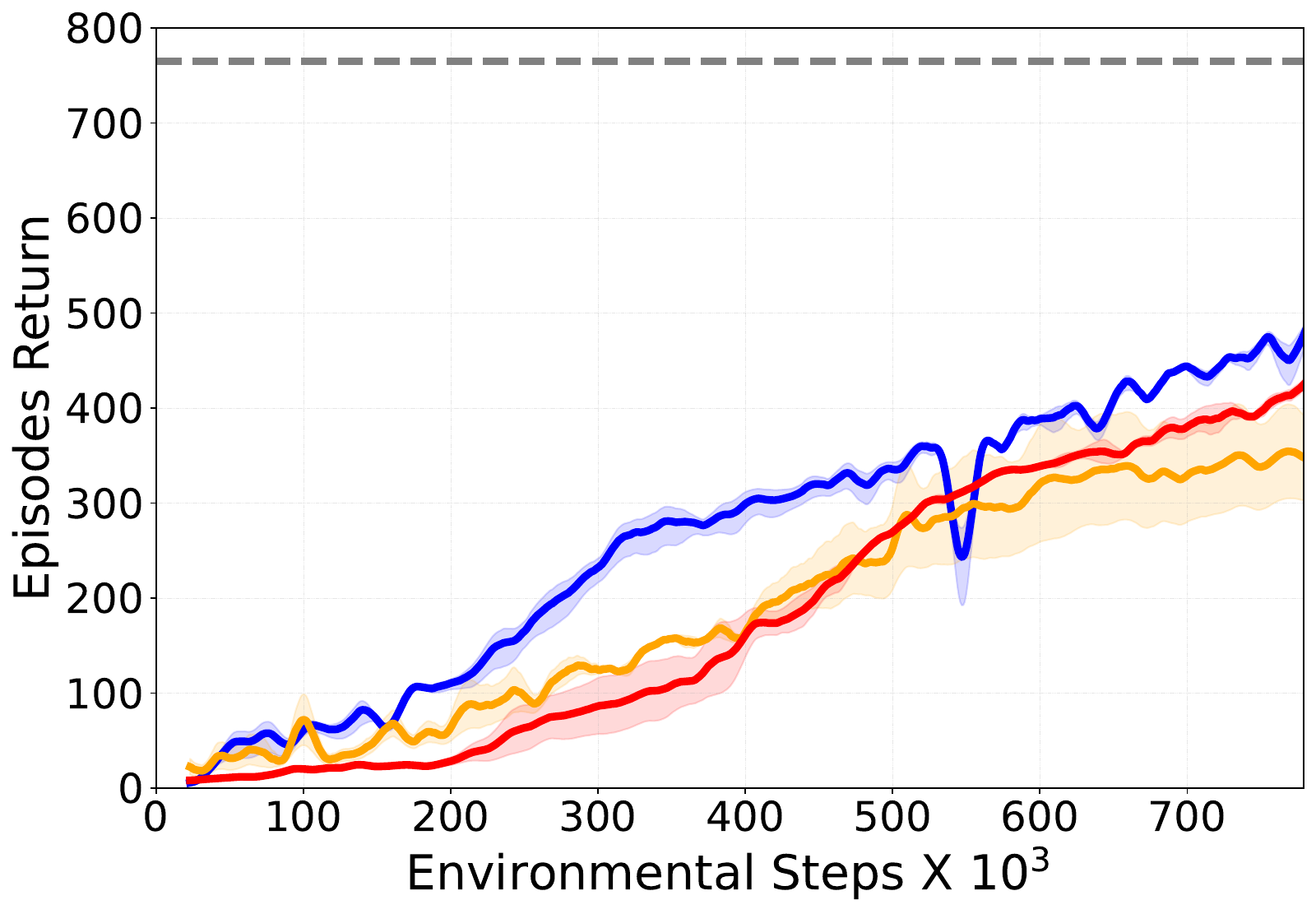}\\
        (b) Visual Cheetah (Pixel-based)
    \end{minipage}
    \includegraphics[width=0.6\linewidth]{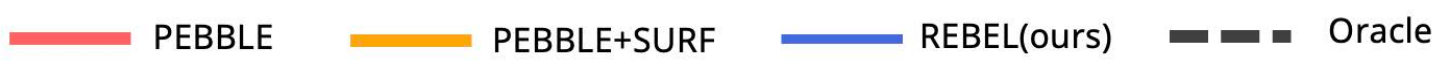}
    \caption{Learning curves on locomotion tasks as measured on the ground truth reward. The solid line and shaded regions, respectively, represent the mean and standard deviation across three runs.}
    \label{fig:combined22}
\end{figure*}

\subsection{Proposed \texttt{REBEL} Algorithm}
\label{proposed}
The main objective of RRLHF is to design a reward function that aligns with human preferences, but we point out that eventually, our designed reward should also be suitable for learning a policy for the downstream task. Our bilevel reformulation (lower-bound) as described in the equation \eqref{bilevel_rlhf_fo_lb} inherently makes the reward learning in \eqref{mle} dependent on the policy as well, through the novel regularization term called \emph{agent preference} which is nothing but the value function at the policy $\pi_{\theta_k^*}$, hence \eqref{mle} would become
    \begin{align}\label{mle_new}
        \min_{\nu} &\underbrace{\mathbb{E}_{\tau_i, \tau_j \sim \mathcal{D}} [y_i \log P_{\nu}(\tau_i > \tau_j) + y_j \log P_{\nu}(\tau_i < \tau_j)]}_{\textbf{Human preference}} 
        \nonumber
        \\
        &+\lambda\underbrace{\mathbb{E}\left[\sum_{h=0}^{H-1} \gamma^h r_{\nu}(s_h, a_h)~|, s_0=s, a_h\sim \pi_{\theta_k^*}\right]}_{\textbf{Agent preference (proposed)}} , 
    \end{align}
    \noindent where we write the objective as a minimization problem and  $\lambda>0$ is the tuning parameter to control the importance of agent preference versus human preference. Intuitively, the additional regularization term would make sure to learn reward which results in a higher value of reward return at the current policy, which implies that it is not very different from the previous reward at which the optimal policy was learned. By regularizing the reward function appropriately, we develop a way to incorporate the policy preference in the learning paradigm, missing from earlier research \cite{park2022surf, lee2021pebble} resulting in sub-optimal alignment.


\section{Experiments}\label{sec:experiments}

\noindent \textbf{Environment:} We consider robot locomotion environments from the DeepMind (DM) Control suite \cite{tassa2018deepmind} with high dimensional visual DM control environments \cite{tassa2018deepmind} to perform detailed experimental evaluations as shown in Figure 2. Specifically, we use the \textit{walker} and the \textit{Cheetah} environments. The objective in each task differs, and the true reward function is typically a function of several attributes. For example, in the \textit{standing} task, the true (hidden) reward incorporates terms that promote an upright posture and maintain a minimum torso height. For the \textit{walking} and \textit{running} tasks, an additional element in the reward function encourages forward movement. Similarly, for Cheetah, the objective of the agent is to run fast up to a certain speed threshold, with rewards based on forward velocity.  We consider two state-of-the-art baselines for preference-based RL, which are PEBBLE \cite{lee2021pebble} and PEBBLE+SURF \cite{park2022surf}. PEBBLE+SURF utilizes data augmentation to improve the performance of PEBBLE,

\vspace{2mm}
\noindent \textbf{Evaluation Metric:} To evaluate the performance of the different methods, we select episodic reward return as a valid metric. The eventual goal of any RL agent is to maximize the expected value of the episodic reward return and this is widely used in the literature. All the hyperparameter details are provided in the supplementary material \cite{appendix}.

\vspace{2mm}
\noindent \textbf{Human Feedback:} Although it would be ideal to evaluate the real-world effectiveness of our algorithm based on actual human feedback, but for simulation and comparisons on benchmark it is hard to collect a large amount of human feedback. Hence, to emulate human feedback, we leverage simulated human teachers as used in prior research \cite{lee2021pebble, park2022surf}, whose preferences are based on ground-truth reward functions which help us to evaluate the agent efficiently. In order to design more human-like teachers, various humanly behaviors like stochasticity, myopic behavior, mistakes etc. are integrated while generating preferences as in \cite{lee2021pebble, park2022surf}.

\vspace{2mm}
\noindent \textbf{Results and Discussions:} We start by plotting the episodic (true and hidden) reward returns for various algorithms in Figure 2. We note that the oracle reward is denoted by a dotted line in all the plots, which is the maximum value of the reward return. All the algorithms would achieve a reward return less than the maximum value. We note that PEBBLE achieves a sub-optimal performance, whereas PEBBLE+SURF (green curve) can achieve better performance, but it requires data augmentation. On the other hand, the proposed {\ours} achieves the best performance compared to all baselines in a sample-efficient manner, in very high-dimensional visual environment, which is closer to the optimal reward. This gain in performance is coming because of the additional agent preference term we add as regularization in the proposed approach (cf. \eqref{mle}).

\section{Conclusion, Limitations, and Future Work
} \label{sec:conclusion}
Sparse reward functions in robotics are ubiquitous and challenging to handle, which usually leads to labor-intensive reward design and reward hacking risks. An interesting existing approach is preference-based learning which utilizes human feedback to infer reward functions, but it faces distribution shift issues. In this work, we addressed these concerns by proposing a novel regularization term called \emph{agent preference} for preference-based learning in robotics control.  We have shown the advantages of the proposed regularization term in experiments. However, limitations include subjective preferences and potential hyperparameter sensitivity. Future research may explore automated preference generation and advanced regularization methods. 

\bibliography{aaai25}
	
\end{document}